\documentclass[10pt, a4paper]{article}

\usepackage[final]{lrec2026} 

\usepackage[acronym, nomain]{glossaries}

\usepackage{caption}
\usepackage{subcaption}

\usepackage{amssymb}
\usepackage{graphicx}

\usepackage[margin=1in]{geometry}
\usepackage{booktabs}
\usepackage{multirow}
\usepackage{float}
\usepackage{array}

\usepackage{xcolor}
\usepackage{tcolorbox}
\tcbuselibrary{breakable}

\definecolor{systembg}{HTML}{F0F0F0}
\definecolor{userbg}{HTML}{E8F4FD}
\definecolor{framecolor}{HTML}{BBBBBB}

\usepackage{tabularx}
\newcolumntype{L}{>{\raggedright\arraybackslash}X}
\newcolumntype{C}{>{\centering\arraybackslash}X}
\newcolumntype{R}{>{\raggedleft\arraybackslash}X}

\usepackage{gb4e}
\noautomath
\newcolumntype{Y}{>{\raggedright\arraybackslash}X}
\usepackage{soul}
\usepackage{xcolor}     

\newcommand{\Tag}[1]{%
  \begingroup
    \setlength{\fboxsep}{2pt}
    \setlength{\fboxrule}{0pt}
    \colorbox{white}{\scriptsize\textbf{#1}}%
  \endgroup
}

\newcommand{\PER}[2]{
  \begingroup
    \setlength{\fboxsep}{2pt}%
    \colorbox[HTML]{9EA5FF}{%
      \strut#1%
      \kern 1pt                        
      \hspace{0.2mm}
      \Tag{#2}%
      \kern 1pt
    }%
  \endgroup
}
\newcommand{\PERP}[2]{
  \begingroup
    \setlength{\fboxsep}{2pt}%
    \colorbox[HTML]{9EA5FF}{%
      \strut#1%
    }%
  \endgroup
}
\newcommand{\ORG}[2]{
  \begingroup
    \setlength{\fboxsep}{2pt}%
    \colorbox[HTML]{C3EDC0}{%
      \strut#1%
      \kern 1pt                        
      \hspace{0.2mm}
      \Tag{#2}%
      \kern 1pt
    }%
  \endgroup
}

\newcommand{\DOC}[2]{
  \begingroup
    \setlength{\fboxsep}{2pt}%
    \colorbox[HTML]{FFC069}{%
      \strut#1%
      \kern 1pt                        
      \hspace{0.2mm}
      \Tag{#2}%
      \kern 1pt
    }%
  \endgroup
}
\newcommand{\DOCP}[2]{
  \begingroup
    \setlength{\fboxsep}{2pt}%
    \colorbox[HTML]{FFC069}{%
      \strut#1%
    }%
  \endgroup
}
\newcommand{\ABS}[2]{
  \begingroup
    \setlength{\fboxsep}{2pt}%
    \colorbox[HTML]{DDB2E1}{%
      \strut#1%
      \kern 1pt                        
      \hspace{0.2mm}
      \Tag{#2}%
      \kern 1pt
    }%
  \endgroup
}
\newcommand{\ABSP}[2]{
  \begingroup
    \setlength{\fboxsep}{2pt}%
    \colorbox[HTML]{DDB2E1}{%
      \strut#1%
    }%
  \endgroup
}
\newcommand{\ACT}[2]{
  \begingroup
    \setlength{\fboxsep}{2pt}%
    \colorbox[HTML]{D88C88}{%
      \strut#1%
      \kern 1pt                        
      \hspace{0.2mm}
      \Tag{#2}%
      \kern 1pt
    }%
  \endgroup
}
\newcommand{\ACTP}[2]{
  \begingroup
    \setlength{\fboxsep}{2pt}%
    \colorbox[HTML]{D88C88}{%
      \strut#1%
    }%
  \endgroup
}
\newcommand{\CLS}[2]{
  \begingroup
    \setlength{\fboxsep}{2pt}%
    \colorbox[HTML]{C2E7E4}{%
      \strut#1%
      \kern 1pt                        
      \hspace{0.2mm}
      \Tag{#2}%
      \kern 1pt
    }%
  \endgroup
}

\glsdisablehyper

\newacronym{llm}{LLM}{Large Language Model}
\newacronym{gelato}{GELATO}{Government, Executive, Legislative, and Treaty Ontology}
\newacronym{lkif}{LKIF}{Legal Knowledge Interchange Format}
\newacronym{ner}{NER}{Named entity recognition}

\title{The GELATO Dataset for Legislative NER}

\name{Matthew Flynn\textsuperscript{1}, Timothy Obiso\textsuperscript{1}, Sam Newman\textsuperscript{1,*}\thanks{*Independent researcher. Work completed while at Brandeis University.}}

\address{\textsuperscript{1}Brandeis University \\
         \{matthewflynn, timothyobiso\}@brandeis.edu, snewman.aa@gmail.com}

\abstract{
This paper introduces GELATO (Government, Executive, Legislative, and Treaty Ontology), a dataset of U.S. House and Senate bills from the 118\textsuperscript{th} Congress annotated using a novel two-level named entity recognition ontology designed for U.S. legislative texts.  We fine-tune transformer-based models (BERT, RoBERTa) of different architectures and sizes on this dataset for first-level prediction. We then use LLMs with optimized prompts to complete the second level prediction. The strong performance of RoBERTa and relatively weak performance of BERT models, as well as the application of LLMs as second-level predictors, support future research in legislative NER or downstream tasks using these model combinations as extraction tools. 
 \\ \newline \Keywords{NER, Dataset, Ontology} }

\begin{document}

\maketitleabstract

\section{Introduction}

The automatic extraction of named entities in the legislative domain is important for supporting various types of research and more advanced NLP methods and applications. Most work done in NER has focused on the more general newswire domain; however, texts from other domains provide an interesting challenge due to differences in mention or text density, specialized terminology, and necessary real-world knowledge.

To this end, we introduce the \gls{gelato}, a two-level NER ontology designed for U.S. congressional legislation. We release an annotated dataset of 131 U.S. House and Senate bills using the \gls{gelato} tag set. Our work provides the baseline for an automatic entity extraction pipeline as new bills are released in the U.S. Congress.

We fine-tune and release several state-of-the-art transformer-based models on the \gls{gelato} dataset for first-level predictions. These predictions are then forwarded with their context to LLMs for second-level prediction, and we release optimized modules for this step. We compute standard NER performance metrics (F1) for both level-one and level-two predictions on our test data.

Our contributions are as follows:
\begin{itemize}
    \item We introduce \gls{gelato}, a two-level NER ontology
    \item We release an annotated dataset using \gls{gelato}
    \item We fine-tune transformer models on \gls{gelato} data for first-level prediction and to provide baselines for future improvement
    \item We propose methods using LLMs to predict second-level NER labels
    \item We release optimized modules for second-level prediction with LLMs
\end{itemize}

All code and data is publicly available on our Github\footnote{\href{https://github.com/Wollaston/gelato}{https://github.com/Wollaston/gelato}}.

\section{Related Work}

Previous research in natural language processing on documents in the legal domain includes domain-specific datasets, benchmarks, and models, many of which are summarized by \citet{küçük2025computationallawdatasetsbenchmarks}. Of particular note is \gls{lkif} \citet{hoekstra2007lkif, hoekstra2009lkif}, an extension of the Knowledge Interchange Format (KIF), a computer language for interacting with knowledge bases designed for the legal domain. This ontology can be used with relational logic systems to represent, compare, and contrast legal systems. LKIF provides a rich framework for representing legal knowledge broadly, though it was not designed with NER annotation in mind. GELATO draws on LKIF's conceptual structure but adapts it into a two-level annotation scheme suitable for sequence labeling.

While NER in the legislative domain is under-researched, much work has been done in the adjacent legal domain. \citet{cardellino2017low, cardellino-etal-2017-legal} create a high-level ontology as a superclass to LKIF, which is a superclass to YAGO \citep{suchanek2007yago, suchanek2024yago}. There are also NER datasets for the legal domain in American English \citep{au-etal-2022-e}, Brazilian Portuguese \citep{luz_etal_propor2018}, German \citep{leitner-etal-2020-dataset}, Indian English \citep{kalamkar-etal-2022-named}, and Chinese \citep{dai-etal-2025-laiw}. These datasets primarily address court judgments and case law, and their ontologies reflect this focus, including Location and Time/Date classes relevant to judicial proceedings. By contrast, GELATO targets congressional legislation and introduces categories for legislative constructs such as Acts, Funds, Programs, and Protected Classes. 

Outside of the legal domain, the 2002 and 2003 CoNLL shared-tasks \citep{tjong-kim-sang-2002-introduction, tjong-kim-sang-de-meulder-2003-introduction} are standard datasets for NER evaluation in Spanish, Dutch, and English. CoNLL's flat ontology of Person, Organization, Location, and Miscellaneous has proven effective for newswire text but was not designed for specialized domains where entities like statutory references, government programs, and classes of people are central. GELATO extends this paradigm with a two-level ontology made of six top-level classes and 30 subclasses tailored to U.S. legislative text.

\begin{table}[!ht]
\begin{center}
\begin{tabularx}{\columnwidth}{ccc|c|X}
      \toprule
      \multicolumn{3}{c|}{\textbf{Annotators}} & \textbf{House Bills} & \multicolumn{1}{c}{\textbf{Senate Bills}} \\
      \textbf{1} & \textbf{2} & \textbf{3} & & \\
      \midrule
      \checkmark & \checkmark &            & 141-170 & 169, 174, 176, 183, 187, 193, 196, 198, 199, 200 \\
      \checkmark &            & \checkmark & 171-200 & 102, 104, 107, 109, 112, 117, 121, 126, 128, 143 \\
                 & \checkmark & \checkmark & 100-140 & 144, 145, 148, 158, 160, 163, 164, 165, 167, 168 \\
      \bottomrule
\end{tabularx}
    \caption{Bills annotated by each annotator}
    \label{tab:bills}
\end{center}
\end{table}

\section{Data and Annotation}

We compiled and annotated 131 U.S. House and Senate bills using the \gls{gelato} tag set. The training ($n=80$) and dev ($n=21$) sets are made up of House bills; and the test set contains only Senate bills ($n=30$). All bills were pulled from the 118\textsuperscript{th} Congress with bill numbers 100-200. 

All House bills in the 100–200 range were annotated, along with 30 Senate bills from the same range selected for their optimal length (3,000–9,000 characters). As detailed in Table \ref{tab:bills}, each annotator processed approximately 80 bills: 60 from the House and 20 from the Senate. Pairwise F1 scores (Table \ref{tab:iaa}) demonstrate high inter-annotator agreement (IAA). While some disagreement is expected given the ontology’s complexity, these scores confirm the clarity of our guidelines and the stability of the underlying framework.

\subsection{Dataset}
The dataset consists of legislative bill text from the House of Representatives and the Senate of the 118\textsuperscript{th} Congress. 
All data was queried from the Congress.gov API\footnote{\href{https://api.congress.gov}{api.congress.gov}} and tokenized. As legislative text is punctuation-dense due to the inclusion of many citations, we split each punctuation mark into its own token.

The training set consists of House bills ($n = 80$) with a total of 77,372 tokens and a mean of 967 tokens per bill. The longest bill contains a total of 10,653 tokens, while the shortest contains 240.

The dev set also consists of House bills ($n = 21$) with a total of 23,819 tokens and a mean of 1,134 tokens per bill. The longest bill contains a total of 3,859 tokens, while the shortest contains 205.

The test set contains Senate bills ($n=30$) of more consistent length with a total of 31,740 and a mean of about 1,058 tokens per bill. The longest Senate bill contains 2,005, while the shortest contains 641. 

Table \ref{tab:gelato_summary} summarizes the \gls{gelato} dataset. Table \ref{tab:counts} provides mention counts for both levels. These counts were obtained via SeqScore \citep{palen-michel-etal-2021-seqscore, lignos-etal-2023-improving}, an evaluation toolkit for sequence labeling tasks. Additionally, all tags and label transitions were validated using SeqScore. 

\begin{table}[t]
    \centering
    \begin{tabular}{ccc|cc}
        \toprule
        \multicolumn{3}{c|}{\textbf{Annotators}} & \textbf{IO F1} & \textbf{Level 1 F1} \\ 
        \textbf{1} & \textbf{2} & \textbf{3} & & \\
        \midrule
        \checkmark & \checkmark &            & 66.35 & 62.53 \\
        \checkmark &            & \checkmark & 77.27 & 73.50 \\
                   & \checkmark & \checkmark & 80.09 & 79.68 \\
        \bottomrule
    \end{tabular}
    \caption{Inter-Annotator Agreement (F1)}
    \label{tab:iaa}
\end{table}

\begin{table*}[t]
\centering

\begin{tabularx}{\textwidth}{L|C|C|C|C}

\toprule
    & \textbf{Train} & \textbf{Dev} & \textbf{Test}  & \textbf{Total}  \\

\midrule
    Bills   & 80 & 21 & 30 & 131 \\
    Tokens  & 77,372 & 23,819 & 31,740 & 132,931 \\

\midrule
    Unique Mentions  & 1,108 & 448 & 556 & 2,112 \\
    Total Mentions   & 3,388 & 1,029 & 1,393 & 5,810 \\

\bottomrule
\end{tabularx}

    \caption{GELATO Summary}
    \label{tab:gelato_summary}
\end{table*}

\subsection{Ontology}

The \gls{gelato} entity type ontology has two levels. The top level has six tags: \textsc{Person}, \textsc{Organization}, \textsc{Document}, \textsc{Act}, \textsc{Abstraction}, and \textsc{Class}. \textsc{Person} and \textsc{Organization} are borrowed from the prototypical CoNLL ontology, though their subclasses are significantly informed by the legislative domain. The other four classes were largely derived from analyzing sample bills through the lens of the \gls{lkif} Ontology. All subclasses are informed by the clustering evaluation of these top-level annotations.

\subsubsection{Person}
The top-level \textsc{Person} class has three subclasses based on the bill domain: \textsc{Member} \textbf{MBC}, \textsc{Title} \textbf{TTL}, and \textsc{Individual} \textbf{IND}. A \textsc{Member} is a member of Congress and is almost always listed as the one introducing a bill or sponsoring a bill. They are present in every bill, appear in a distinct context, and serve a downstream interest. A \textsc{Title} is a mention of a person in a specific position, such as \textit{Speaker} or \textit{President}. An \textsc{Individual} is a \textsc{Person} mention that falls into neither of the above categories.

\begin{exe}
    \ex January 9, 2023 Mr. \PER{Biggs}{MBC} introduced the following bill
    \ex for a period to be subsequently determined by the \PER{Speaker}{TTL}
\end{exe}



 







\begin{table*}[t]
\centering
\begin{tabularx}{\textwidth}{l l|c@{\hskip 3pt}c@{\hskip 3pt}c@{\hskip 3pt}c@{\hskip 3pt}c|X}
\toprule
\textbf{Ontology} & \textbf{Locale} & \rotatebox{60}{\textbf{Person}} & \rotatebox{60}{\textbf{Organization}} & \rotatebox{60}{\textbf{Location}} & \rotatebox{60}{\textbf{Legal Text}} & \rotatebox{60}{\textbf{Time/Date}} & \textbf{Other} \\
\midrule
GELATO & American English &
  \checkmark & \checkmark & & \checkmark & & Abstraction, Act, Class \\
Cardellino \emph{et al.} & American English &
  \checkmark & \checkmark & & \checkmark & & Abstraction, Act \\
Leitner \emph{et al.} & German &
  \checkmark & \checkmark & \checkmark & \checkmark & & Case‐by‐case Regulation \\
Kalamkar \emph{et al.} & Indian English &
  \checkmark & \checkmark & \checkmark & \checkmark & \checkmark & Court, Case Number \\
LeNER-Br & Brazilian Portuguese &
  \checkmark & \checkmark & \checkmark & \checkmark & \checkmark & \\
E-NER  & American English &
  \checkmark & \checkmark & \checkmark & \checkmark & & Court, Miscellaneous \\
\bottomrule
\end{tabularx}
\caption{Comparison of legal‐domain NER ontologies across different papers and languages}
\label{tab:ner‐ontologies}
\end{table*}

\subsubsection{Organization}
The top level \textsc{Organization} class has eight subclasses based on the bill domain: \textsc{Nation} \textbf{NAT}, \textsc{State} \textbf{STA}, \textsc{Locality} \textbf{LOC}, \textsc{Committee} \textbf{COM}, \textsc{Agency} \textbf{AGC}, \textsc{Legislative Body} \textbf{LEG}, \textsc{International Institution} \textbf{INT}, and \textsc{Association} \textbf{ASC}. 
\textsc{Nation}, \textsc{State}, and \textsc{Locality} are the three levels of geopolitical entities covered in the \textsc{Org} class. The \textsc{Nation} class is named as such rather than \textit{country} to avoid issues of official UN recognition and to include American Indian nations, etc. (\ref{state}) is an example of a state. 

The classes \textsc{Legislative Body} and \textsc{Committee} cover a fairly static set of entities, where \textbf{LEG} mostly encompasses $\{\textit{Senate, House of Representatives, Congress}\}$ and \textbf{COM} covers their committees, as demonstrated in (\ref{legnat}) and (\ref{com}).
\textsc{Agency} and \textsc{International Institution} encompass mentions of U.S. government agencies, (\textit{departments, bureaus, administrations,} etc.) and mentions of international organizations like the \textit{United Nations}, respectively.
\textsc{Association} is the \textbf{ORG} class's catch-all subclass for capturing mentions like \textit{Planned Parenthood}, etc.

\begin{exe}
    \ex \label{state} Mr. \PER{Biggs}{MBC} , Mr. \PER{Nehls}{MBC} , Mrs. \PER{Miller}{MBC} of \ORG{Illinois}{STA}
    \ex \label{legnat} Be it enacted by the \ORG{Senate}{LEG} and \ORG{House of Representatives}{LEG} of the \ORG{United States of America}{NAT}
    \ex \label{com} which was referred to the \ORG{Committee on Ways and Means}{COM}
\end{exe}

\subsubsection{Document}
The \textsc{Document} class takes more inspiration from the \gls{lkif} ontology and its mereological and dependency relationships. It encompasses those physical entities on which \textsc{Act}s depend.
The top level \textsc{Document} class has six subclasses based on the bill domain: \textsc{Code} \textbf{CODE}, \textsc{Bill} \textbf{BILL}, \textsc{Reference} \textbf{REF}, \textsc{Parenthetical} \textbf{PAR}, \textsc{Treaty} \textbf{TRE}, and \textsc{Report} \textbf{REP}.
\textsc{Code} covers the foundational documents of United States law. This includes documents such as the \textit{United States Code}, which is frequently referenced and amended in legislation.
The \textsc{Bill} class exists almost exclusively for capturing the extremely consistent mentions of bill numbers, as demonstrated in (\ref{bill}).

The \textsc{Reference} and \textsc{Parenthetical} classes capture the much more varied mentions of specific \textit{titles}, \textit{sections}, \textit{paragraphs}, etc. that are directly referenced and amended in most bills, as demonstrated in (\ref{ref}).
The \textsc{Report} class is the somewhat catch-all subclass of \textsc{Document}, and includes mentions such as \textit{President's Budget} and similar documents.
The \textsc{Treaty} class is for treaty mentions, which did not show up in our dataset. We include this subclass in the \gls{gelato} ontology because treaties don't fit the concept of \textsc{Report}, which isn't a true catch-all, and are important to capture, though rare.

\begin{exe}
    \ex \label{bill} [\DOC{H.R. 189}{BILL} Introduced in House (IH)]
    \ex \label{ref} statement pursuant to \DOCP{section 102 of the}{}
    
    \DOCP{National Environmental Policy Act of}{} 
    
    \DOC{1969}{REF} \DOC{(42 U.S.C. 4332)}{PAR}
\end{exe}

\subsubsection{Act}
The \textsc{Act} class depends on the \textsc{Document} class, and encompasses two well-defined subclasses: \textsc{Public Act} \textbf{PA} and \textsc{Amendment} \textbf{AMD}.
\textsc{Public Act} mentions are those that refer to an act by name, and \textsc{Amendment} mentions are those that refer to amendments to the constitution.

\begin{exe}
    \ex prior to the enactment of the \ACT{American Rescue Plan Act}{PA}
    \ex as guaranteed by the \ACTP{Second}{} \ACT{Amendment to the Constitution}{AMD}
\end{exe}

\subsubsection{Abstraction}
The \textsc{Abstraction} class has 9 subclasses: 8 based on the legislative domain and a \textsc{Misc} subclass. The general goal of \textsc{Abstraction} is to capture entities that are the result of legislative action. The 8 non-miscellaneous subclasses are: \textsc{Program} \textbf{PRO}, \textsc{Session} \textbf{SES}, \textsc{System} \textbf{SYS}, \textsc{Infrastructure} \textbf{INF}, \textsc{Fund} \textbf{FUND}, \textsc{Doctrine} \textbf{DTR}, \textsc{Specification} \textbf{SPC}, and \textsc{Case} \textbf{CASE}.

The \textsc{Program} subclass encompasses programs established and managed by the government, such as \textit{Medicare} and \textit{Social Security}.
The \textsc{System} and \textsc{Infrastructure} subclasses cover the frameworks established and maintained by the government, often in support of a \textsc{Program}. The distinction lies in whether the entity is tangible, such as \textit{International Outfall Interceptor}, for the latter, or not, such as the \textit{National Forest System} in (\ref{nfs}), or a database or website for the former.
Mentions of the \textsc{Fund} subclass include those referring to the \textit{General Fund of the Treasury} or other specific funds such as those in (\ref{fund}).
The \textsc{Doctrine} subclass covers somewhat abstract policies, principles, and rights, such as \textit{Free Speech} or the \textit{Fairness Doctrine} in (\ref{fair}).
The \textsc{Case} subclass, like the \textsc{Treaty} subclass, is unpopulated in our dataset, but encompasses mentions of court cases such as \textit{McCulloch v. Maryland}, which are referenced in bills when they establish relevant legal precedents or support the need for legislative changes.
The \textsc{Specification} subclass covers mentions of specific, named legal specifications and classifications like \textit{Schedule I} and \textit{340b} in (\ref{340b}).

\begin{exe}
    \ex \label{nfs} forest management activity conducted on \ABS{National Forest System}{SYS} land
    \ex \label{record} maintained in the \ABSP{National Firearms}{} \ABS{Registration and Transfer Record}{SYS}
    \ex \label{fund} held by the \ABSP{Old-Age and Survivors}{} \ABS{Insurance Trust Fund}{FUND}
    \ex \label{fair} present opposing viewpoints on controversial issues of public importance, commonly referred to as the `\ABS{Fairness Doctrine}{DTR}'
    \ex \label{340b} subject to an agreement under \DOCP{section 340B of the Public Health Service}{} \DOC{Act}{REF} shall include the \ABS{340B}{SPC} modifier established by the \PER{Secretary}{TTL}
\end{exe}

\subsubsection{Class}
The \textsc{Class} class refers to a class of people. The initial motivator for including the class was capturing mentions of protected classes qualified for special protection by law, i.e. $\{$\textit{Age, Ancestry, Color, Disability, Ethnicity, Gender, HIV/AIDS status, Military Status, National Origin, Pregnancy, Protected Veteran Status, Race, Religion, Sex, Sexual Orientation}$\}$. For these mentions we created the subclass \textsc{Protected Class} \textbf{PC}. When annotating the pilot, we found instances where other classes of people were a subject of a bill and consistently referenced throughout (\ref{npc}), and we decided to add the subclass \textsc{Non-Protected Class} \textbf{NPC} to capture their mentions.

We also started noticing repeated spans like \textit{State} and \textit{Committee} referenced throughout bills, leading us to recognize that \gls{gelato} lacked a type for classes of organization, which would complete the logical parallel (i.e.
$\textbf{PER}:\textbf{CLS}::\textbf{ORG}:\rule{.8cm}{0.15mm}$). However, we ultimately decided that the mentions that would fall within the class were not worth capturing, as their presence in a bill did not meaningfully differentiate that bill from another.

\begin{exe}
    \ex processing of claims for temporary disability ratings for \CLS{veterans}{PC} described
    \ex \label{npc} submit to \ORG{Congress}{LEG} a recommendation on the number of \CLS{refugees}{NPC} who may be admitted
\end{exe}

Table \ref{tab:ner‐ontologies} shows a comparison between \gls{gelato} and other ontologies for adjacent domains.

\subsection{Annotation}
The two levels of labels for each document and mention in \gls{gelato} were completed in two steps with three total annotators. 

The first level of \gls{gelato} annotation for each document was completed by two annotators using Label Studio \citep{label-studio}. All annotators shared the same configuration and Label Studio instance. After annotation, all three annotators participated in the adjudication process. All tokens with differing tags were discussed by the group of annotators and resolved through group consensus. After adjudication, the annotators proofread these adjudicated tags to obtain the final gold-standard first-level data, ensuring quality and consistency.

To obtain the second-level \gls{gelato} labels, the three annotators reviewed the gold labels together. For each gold label mention, the annotators decided which second-level label of the respective first-level label was most appropriate, and tagged it accordingly. Together, this two step process resulted in the complete and fully adjudicated two-level, gold-standard \gls{gelato} dataset.  

\section{Multi-level NER with LLMs}
We propose a no-training approach to level-two NER label prediction that evokes the generalized knowledge LLMs have obtained during pretraining through optimized prompts that include specific instructions. There are many challenges when performing NER with LLMs, and there have been many ways proposed of doing so \cite{labrak-etal-2024-zero, subedi-etal-2024-exploring, wang-etal-2025-gpt}.

In this paper, we use DSPy \citep{khattab2022demonstrate, khattab2024dspy} to find optimized prompts. The base prompt for the LLM is generated with its level-one label, a context window of 50 tokens on each side (totalling 100), and the possible level-two labels. This includes input and output field descriptions, as well as a template structure to frame the response. In the user message, this template structure is filled in with the necessary context before the model is prompted to generate a structured response from which the level-two label can be extracted. We chose 50 tokens for the context window as it is reasonably sized for model performance while limiting the total number of tokens per prompt. With this optimized configuration, we prompt the LLM to classify the provided mention. 

This approach allows us to take advantage of the strengths of LLMs and smaller, fine-tuned transformer language models. We use the smaller model to process every token and classify them according to a simpler ontology. Once there is a level-one label for a mention, we prompt an LLM with few-shot examples to perform level-two classification.

\section{Experiments}

\subsection{Level One}
We fine-tune a series of transformer language models on the \gls{gelato} dataset. These results are summarized in Table \ref{tab:gelato_f1}. Our results show the strengths and weaknesses of models of different sizes and pretraining strategies. We report test results after fine-tuning base and large models of BERT and RoBERTa \citep{devlin-etal-2019-bert, liu2019robertarobustlyoptimizedbert} on \gls{gelato}.

To obtain the optimal hyperparameters for each model, we conducted a sweep of 50 runs on Weights and Biases \citep{wandb} using Bayesian hyperparameter optimization \citep{snoek2012practical} over \texttt{learning\_rate}, \texttt{batch\_size}, \texttt{epochs}, \texttt{weight\_decay}, and \texttt{warmup\_ratio}. Learning rate was selected from a log uniform distribution from 1\text{\textsc{e-}}5 to 6\text{\textsc{e-}}4; batch size was varied between 1, 2, 4, 8, 16, 32, 64, and 128. Epochs between 1 and 50 were tested. Weight decay and warmup ratio were both varied between 0.0 and 0.5. The best set of hyperparameters for each model is included in Table \ref{tab:gelato_f1}.

\begin{table*}[]
    \centering
    \begin{tabularx}{0.8\textwidth}{L|ccccc|c|c}
        \toprule
        \textbf{Model}      & \textbf{LR} & \textbf{BS} & \textbf{E} & \textbf{WD} & \textbf{WR} & \textbf{Lvl 1 F1} & \textbf{Lvl 2 F1} \\ 
        \midrule
        bert-base-cased     & 3\text{\textsc{e-}}5 & 1  & 44 & 0.4 & 0.5 & 68.09 & 53.26 \\
        bert-large-cased    & 1\text{\textsc{e-}}5 & 1  & 34 & 0.0 & 0.2 & 67.16 & 53.14 \\
        
        roberta-base        & 1\text{\textsc{e-}}4  & 2  & 47 & 0.1 & 0.2 & 84.87 & 64.79 \\
        roberta-large       & 5\text{\textsc{e-}}5 & 1  & 48 & 0.5 & 0.1 & \textbf{88.72} & \textbf{67.44} \\
        \midrule
        Gold Level One Labels & - & - & - & - & - & - & 76.60 \\
        \bottomrule
    \end{tabularx}
    \caption{Best F1 scores and hyperparameter configuration for each model; LR = learning rate, BS = batch size, E = epochs, WD = weight decay, WR = warmup ratio}
    \label{tab:gelato_f1}
\end{table*}

\subsection{Level Two}
We obtain level-two labels using the labels predicted by the transformer models in level one. To find optimized prompts for each level-one label, we use the MiPROv2 optimizer \citep{opsahl-ong-etal-2024-optimizing} from DSPy with Qwen3/Qwen-32B \citep{qwen3technicalreport} as the inference model.\footnote{\href{https://huggingface.co/Qwen/Qwen3-32B}{https://huggingface.co/Qwen/Qwen3-32B}}

For the prediction step, we also used Qwen/Qwen3-32B self-hosted with vLLM \citep{kwon2023efficient}. We loaded each optimized DSPy module and routed the prompting for the level-two label based on the level-one prediction. Because of this, if the level-one label was incorrect, the level-two label would be incorrect as well. The predicted level-two label was mapped back to its respective level-one label for alignment and scoring. The level-two results are summarized alongside the level-one results in Table \ref{tab:gelato_f1}, and a detailed breakdown of F1 by label for the best-performing \texttt{roberta-large} is presented in Table \ref{tab:stage2_f1}.

\begin{table*}
    \centering
    \begin{tabular}{llc|llc}
        \toprule
        \textbf{Type} & & \textbf{F1} & \textbf{Type} & & \textbf{F1} \\
        \midrule
         Abstraction & & 67.80              & & Code & 5.26 \\
            & Case & --                      & & Parenthetical & 13.33 \\
            & Doctrine & 0                  & & Reference & 61.54  \\
            & Fund & 66.67                  & & Report & 0 \\
            & Infrastructure & 61.54        & & Treaty & -- \\
            & Misc & 0                  & Organization  & & 90.64 \\
            & Program & 51.06               & & Agency & 88.37 \\
            & Session & 100.00              & & Association & 31.11 \\
            & Specification & --             & & Committee & 91.25 \\
            & System & 28.57                & & International Institution & 85.71 \\            
         Act & & 88.24                      & & Legislative Body & 95.27\\                
            & Amendment & 0                 & & Locality & 50.00 \\           
            & Public Act & 89.31            & & Nation & 83.46 \\          
         Class & & 27.52                    & & State & 69.39 \\                  
            & Non-Protected Class & 9.76 & Person & & 96.65 \\   
            & Protected Class & 14.81       & & Member & 46.84\\
         Document & & 94.99                 & & Name & 0  \\
            & Bill & 95.24                  & & Title & 93.22 \\               
                              
        \bottomrule
    \end{tabular}
    \caption{F1 scores for level-two predictions using roberta-large and Qwen/Qwen3-32B; -- indicates the label was not present}
    \label{tab:stage2_f1}
\end{table*}

The F1 scores for the level-two predictions show a wide variance in performance by level-one and level-two label types. Models perform better with more concrete types that exist in specific contexts in the training data, such as \textsc{Bill} and \textsc{Title}, while struggling with more abstract or referential types, such as \textsc{Parenthetical}. There could be some performance loss as well due to the relative contextual or semantic similarity of certain types, such as \textsc{Member} and \textsc{Name}.

\section{Results}

As expected, the larger RoBERTa models perform better at this task; the best-performing model is \texttt{roberta-large} with an F1 of 88.72. The BERT base-sized model performed the worst with an F1 of 68.09. Because of the timing of the 118\textsuperscript{th} Congress (2023-2025), it is impossible that the exact text of any of these bills was included in the training data. U.S. Congress bills are publicly available so it cannot be concluded that there were no legislative bills or texts in any of the pretraining data; despite this, our results showcase the ability of transformer models to perform domain-aligned named entity extraction in a novel domain. 

Additionally, legislative bills have many unique features such as their structure, format, references, style, etc. Documents of this format may not have been included in significant numbers in the pretraining of the language models we experiment with. As a result, the token sequences we are providing to the models are very unique and likely novel. In addition, we also train with fewer tokens than large-scale general NER datasets such as CoNLL. As such, the evaluation metrics from models trained for our task are lower.

For the level-two prediction task, in which the Qwen model makes fine-grained predictions based off of level-one labels, the F1 score is approximately 20 points lower for RoBERTa models, and approximately 14 to 15 points lower for BERT models, when compared to their level-one F1 counterpart. Due to the cascading nature of level one errors on level two predictions, we also report the performance of the Qwen model conditioned on the gold level one labels.

It is unclear if Qwen/Qwen3-32B's pretraining data included these specific bills or content from the 118\textsuperscript{th} Congress, but it can be reasonably concluded that it has seen similar content. This is generally expected with LLMs and their vast pretraining corpora, and it is this knowledge and generalizability that makes using LLMs an interesting choice for second-level prediction in two-level ontologies.

\section{Error Analysis}

\begin{figure*}
     \centering
     \begin{subfigure}{0.45\textwidth}
         \centering
         \includegraphics[width=\linewidth]{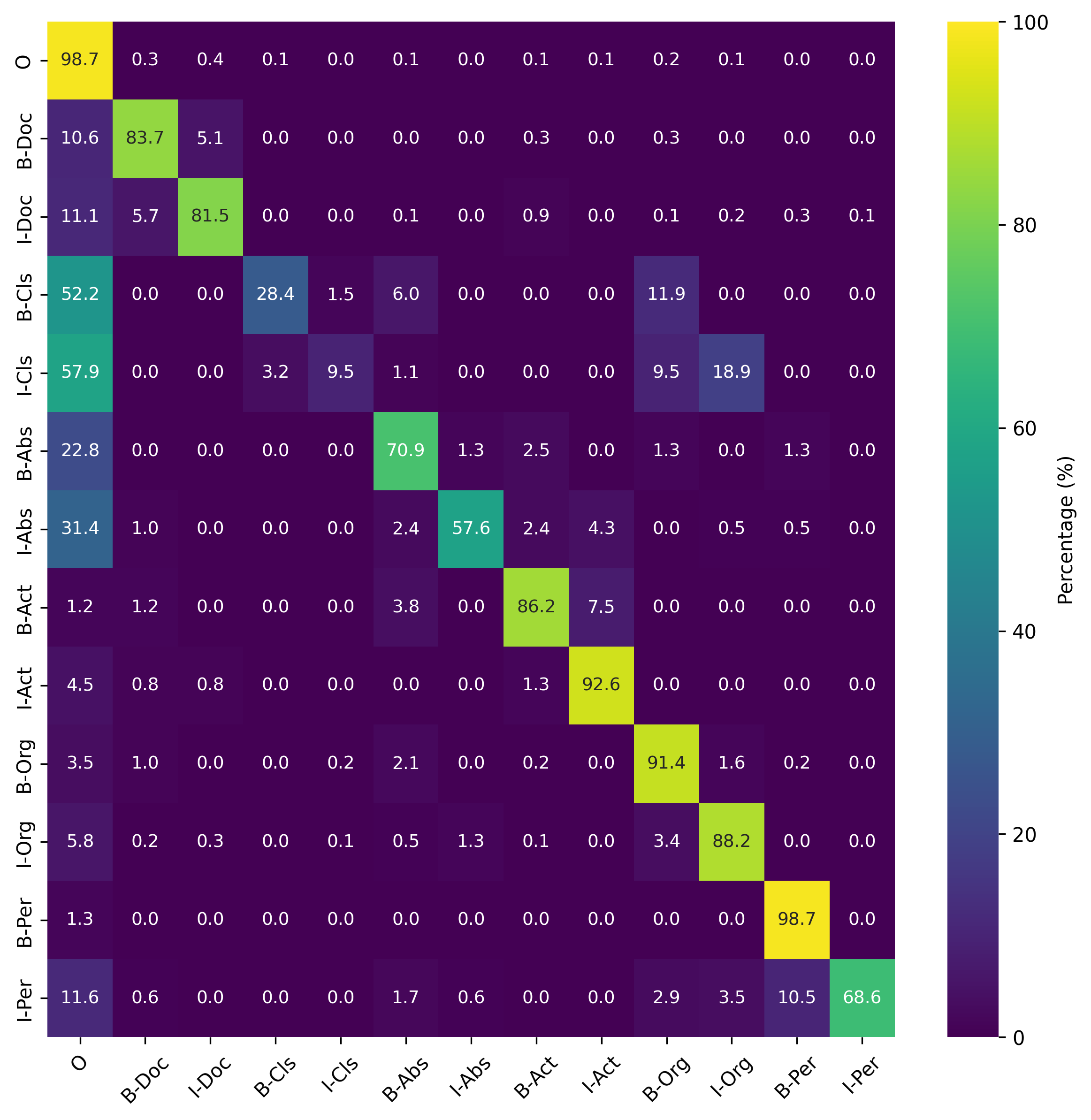}
         \caption{bert-base-cased}
         \label{fig:2a}
     \end{subfigure}
     \begin{subfigure}{0.45\textwidth}
         \centering
         \includegraphics[width=\linewidth]{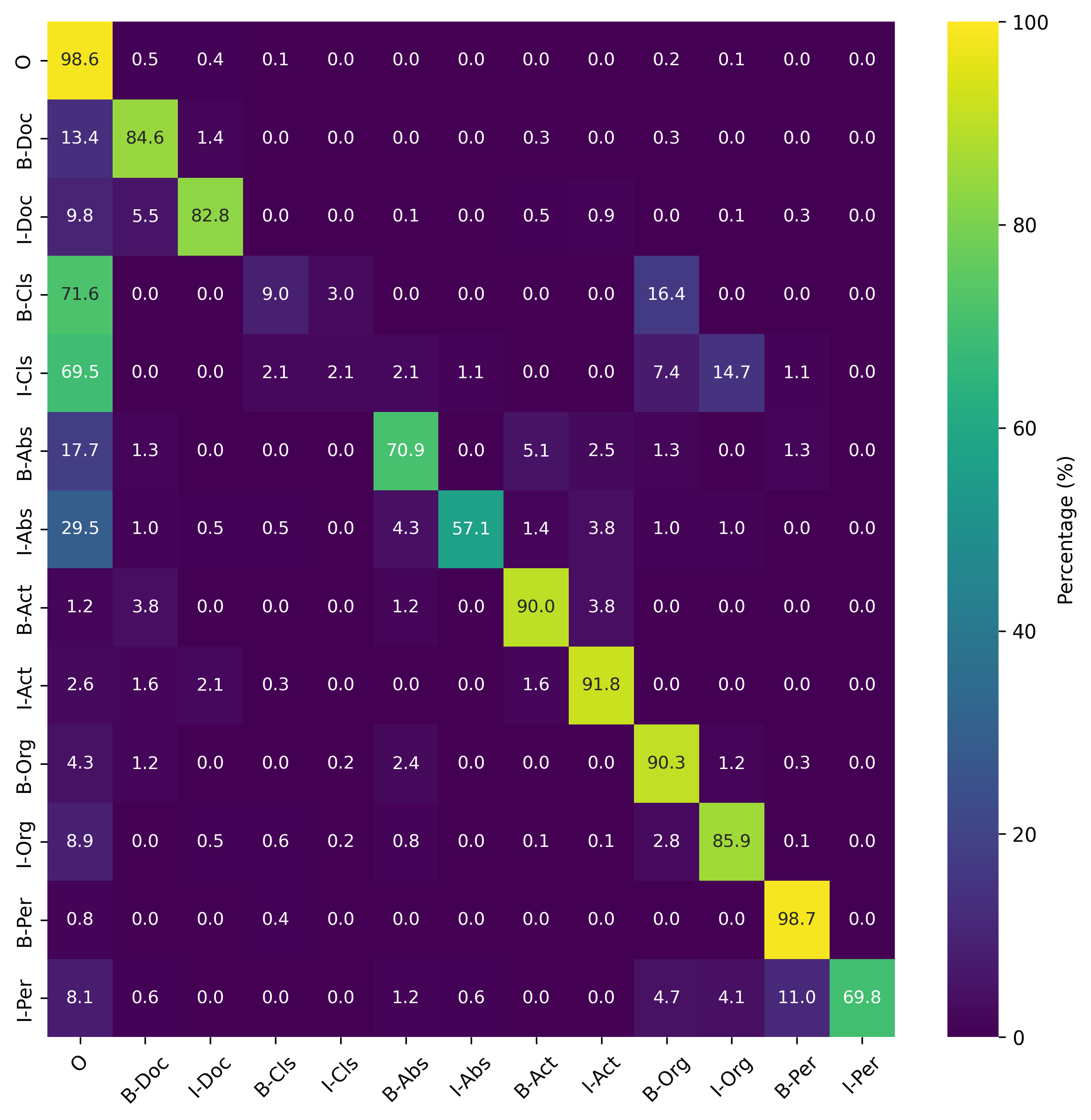}
         \caption{bert-large-cased}
         \label{fig:2b}
     \end{subfigure}
    \begin{subfigure}{0.45\textwidth}
         \centering
         \includegraphics[width=\linewidth]{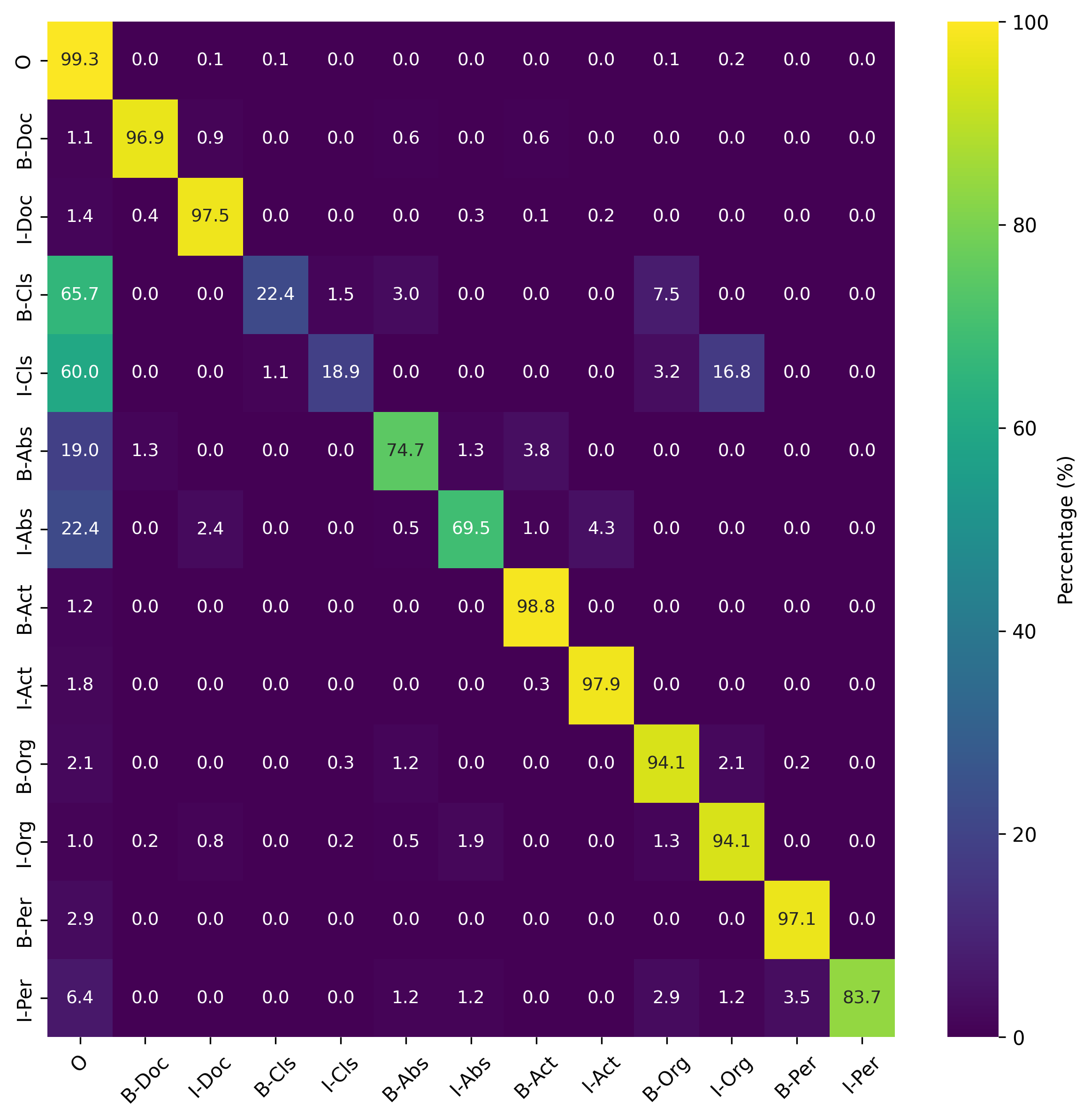}
         \caption{roberta-base}
         \label{fig:2c}
     \end{subfigure}
     \begin{subfigure}{0.45\textwidth}
         \centering
         \includegraphics[width=\linewidth]{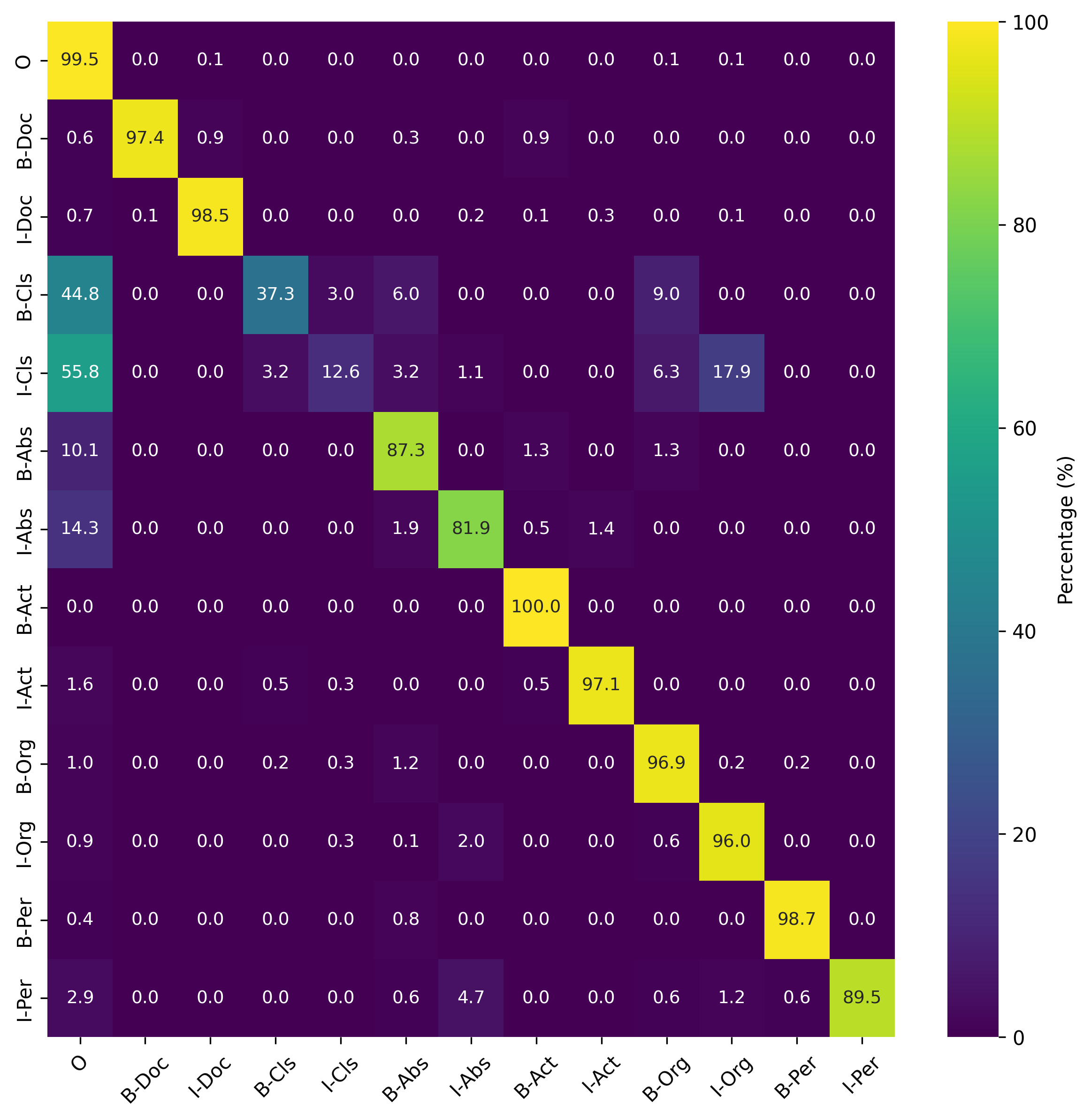}
         \caption{roberta-large}
         \label{fig:2d}
     \end{subfigure}
     \caption{Level one confusion matrices}
     \label{fig:confusion}
\end{figure*}

BERT models performed much worse than RoBERTa models. To investigate why, we first produced confusion matrices to gain better insight into what tags the models struggled with in general, and relative to one another. These are shown in Figure \ref{fig:confusion}.

Of particular note is that all models performed poorly when classifying \textsc{Class} mentions. We believe this is due to the nuance required in order to correctly annotate and tag \textsc{Class} mentions, especially relative to a \textsc{Person} mention. This is likely due to the fact that many \textsc{Class} mentions are plural versions of words that would otherwise be tagged as \textsc{Person} if in their singular forms. 

And while \gls{gelato} distinguishes between \textsc{Person} and \textsc{Class} mentions, which differ primarily in number, \gls{gelato} notably does not make the distinction between \textsc{Organization} and some group of organizations. Perhaps the models would have performed better if this distinction also exists to help better generalize over number.

The models in our experiments also differ in their vocabulary sizes and tokenizers. BERT has a vocabulary size of 30,522 and uses the WordPiece subword tokenization strategy \citep{schuster2012japanese, song2020fast}. In contrast, RoBERTa has a vocabulary size of 50,265 and uses the byte version of Byte-Pair Encoding for tokenization \citep{sennrich2015neural}. We believe that the differences in vocabulary size, as well as the different tokenization strategies, also contribute to the relative performance differences of these models.

As for LLMs, they have vast amounts of general knowledge from their pretraining, and this can impact performance when these general expectations do not align with those of domain-specific interests, like those of \gls{gelato}. For example, LLMs frequently confused the \textsc{Name} subclass of the first-level \textsc{Person} class with the other \textsc{Member} subclass. This can be expected confusion with LLMs as \textsc{Member} is a specific type of name in \gls{gelato}. This suggests that when designing NER ontologies that LLMs will use, it is important to have clear semantic distinction to avoid LLM confusion.

Interestingly, in rare cases, the LLM will refuse to choose one of the provided level-two labels, noting that none of the options are satisfactory in its generated reasoning and leading to an automatic error. This occurred between one and seven times with level-two predictions for each of the four generated level-one prediction sets, and more frequently with level-one labels from the less-performant BERT models.

As well, with the two-level ontology and predictions, any error made in level one would cascade to level two. Therefore, it is expected that level-two F1 scores would lag behind their level-one counterparts, and that level-two predictions generated from RoBERTa predictions would strongly outperform those from BERT models. Any increase in performance for level-one predictions would therefore lead to a likely performance boost for level two as well. This is particularly notable for any misalignment errors in level one, as level two does not perform any BIO-encoding and instead relies on the encoding generated in level one.

\section{Discussion}

Our experiments on \gls{gelato} reveal a number of insights into the challenges present in legislative NLP, particularly NER.

The superior performances of the RoBERTa model emphasizes how a larger vocabulary size and more diverse pretraining data can greatly improve model performance on the same task for a model of a similar architecture.

The complete absence of \textsc{Class} in most model predictions indicates an open challenge in legislative NER. According to the confusion matrices in Figure \ref{fig:confusion}, \textsc{Class} was most commonly predicted to be `O'. The similarity between items in the \textsc{Person} and \textsc{Class} classes are subtle; future work can explore why and how models may need more training examples containing \textsc{Class} to learn to differentiate it from general lexical items and \textsc{Person}.

Based on the fine-tuning results, it seems that transfer learning is not applicable here and misalignment between NER domains leads to decreased performance over a single-task training paradigm. It also seems that there is room for BERT-era models to improve and reach the performance of RoBERTa.

Although not explored in this work, this approach could also be used to perform NER on traditional CoNLL-style, single-level datasets. First, a smaller language model would be trained to not classify, but rather identify named entities. In other words instead of 2n+1 labels for n classes, you would only need three labels (BIO). Our approach would take these identified mentions and prompt an LLM to do the level 1 classification. One shortcoming of this approach is that it is dependent on the performance of a level-one classifier. 

Overall, our results show that pre-trained models offer competitive performance on the \gls{gelato} dataset and can be reliably deployed for downstream tasks, in addition to future research. \gls{gelato} is a useful benchmark for investigating the performance of models at NER in the legislative domain.

\section*{Limitations}

The \gls{gelato} dataset is limited to House and Senate Bills 100-200 from the 118th Congress. This limits our dataset temporally and stylistically. This may hurt the performance of our models on bills from previous or future congressional sessions, from different legislative chambers at the state level, or from other English-speaking countries.

While \gls{gelato} aims to comprehensively cover the most common entities in U.S. legislative documents, Treaty and Case have no mentions across our train, dev, and test sets which limits our ability to use this dataset to extract those entity types.

One limitation of our two-stage prediction pipeline is that errors made in stage one propagate through the pipeline and are irrecoverable in stage two. When performing prompt optimization, we find one optimized prompt for each level-one label, giving us six optimized prompts in total. Other designs of this system may involve a single prompt or even a prompt for each level-two label.
 
\section*{Ethical Considerations}

All bills in the \gls{gelato} dataset are publicly available U.S. government documents obtained via the Congress.gov API and are therefore in the public domain and not subject to copyright restrictions. We release our code and models under a permissive open-source license to encourage more research with our models and data. 

Three graduate student annotators (the authors) with training in linguistics and NLP collaboratively created \gls{gelato} through a two-stage process with full adjudication of any disagreements. This is a descriptive annotation; for example, this ontology includes Protected Class and Non-Protected Class subclasses that are consistent with U.S. anti-discrimination law definitions. 

\gls{gelato} can support beneficial applications including legislative tracking, policy analysis, and government transparency initiatives. However, automated entity extraction could also enable potentially harmful uses such as targeted analysis of how specific groups are referenced in legislation or identification of individual legislators for inappropriate purposes.

\section{Acknowledgments}
We thank Constantine Lignos for his helpful comments and insightful feedback on this paper. We also thank anonymous reviewers whose feedback helped to improve this paper.

\section{Bibliographical References}\label{sec:reference}

\bibliographystyle{lrec2026-natbib}
\bibliography{gelato}

@article{hoekstra2007lkif,
  title={The lkif core ontology of basic legal concepts.},
  author={Hoekstra, Rinke and Breuker, Joost and Di Bello, Marcello and Boer, Alexander and others},
  journal={LOAIT},
  volume={321},
  pages={43--63},
  year={2007}
}

@incollection{hoekstra2009lkif,
  title={LKIF core: Principled ontology development for the legal domain},
  author={Hoekstra, Rinke and Breuker, Joost and Di Bello, Marcello and Boer, Alexander},
  booktitle={Law, ontologies and the semantic web},
  pages={21--52},
  year={2009},
  publisher={IOS Press}
}

@misc{küçük2025computationallawdatasetsbenchmarks,
      title={Computational Law: Datasets, Benchmarks, and Ontologies}, 
      author={Dilek Küçük and Fazli Can},
      year={2025},
      eprint={2503.04305},
      archivePrefix={arXiv},
      primaryClass={cs.CL},
      url={https://arxiv.org/abs/2503.04305}, 
}

@InProceedings{luz_etal_propor2018,
          author = {Pedro H. {Luz de Araujo} and Te\'{o}filo E. {de Campos} and
                    Renato R. R. {de Oliveira} and Matheus Stauffer and
                    Samuel Couto and Paulo Bermejo},
          title = {{LeNER-Br}: a Dataset for Named Entity Recognition in {Brazilian} Legal Text},
          booktitle = {International Conference on the Computational Processing of Portuguese
                       ({PROPOR})},
	  publisher = {Springer},
	  series = {Lecture Notes on Computer Science ({LNCS})},
	  pages = {313--323},
          year = {2018},
          month = {September 24-26},
          address = {Canela, RS, Brazil},	  
	  doi = {10.1007/978-3-319-99722-3_32},
	  url = {https://teodecampos.github.io/LeNER-Br/},
}

@inproceedings{cardellino2017low,
  title={A low-cost, high-coverage legal named entity recognizer, classifier and linker},
  author={Cardellino, Cristian and Teruel, Milagro and Alemany, Laura Alonso and Villata, Serena},
  booktitle={Proceedings of the 16th edition of the International Conference on Artificial Intelligence and Law},
  pages={9--18},
  year={2017}
}

@inproceedings{suchanek2007yago,
  title={Yago: a core of semantic knowledge},
  author={Suchanek, Fabian M and Kasneci, Gjergji and Weikum, Gerhard},
  booktitle={Proceedings of the 16th international conference on World Wide Web},
  pages={697--706},
  year={2007}
}

@inproceedings{suchanek2024yago,
  title={Yago 4.5: A large and clean knowledge base with a rich taxonomy},
  author={Suchanek, Fabian M and Alam, Mehwish and Bonald, Thomas and Chen, Lihu and Paris, Pierre-Henri and Soria, Jules},
  booktitle={Proceedings of the 47th International ACM SIGIR Conference on Research and Development in Information Retrieval},
  pages={131--140},
  year={2024}
}

@inproceedings{schuster2012japanese,
  title={Japanese and korean voice search},
  author={Schuster, Mike and Nakajima, Kaisuke},
  booktitle={2012 IEEE international conference on acoustics, speech and signal processing (ICASSP)},
  pages={5149--5152},
  year={2012},
  organization={IEEE}
}

@article{song2020fast,
  title={Fast wordpiece tokenization},
  author={Song, Xinying and Salcianu, Alex and Song, Yang and Dopson, Dave and Zhou, Denny},
  journal={arXiv preprint arXiv:2012.15524},
  year={2020}
}

@article{sennrich2015neural,
  title={Neural machine translation of rare words with subword units},
  author={Sennrich, Rico and Haddow, Barry and Birch, Alexandra},
  journal={arXiv preprint arXiv:1508.07909},
  year={2015}
}

@article{snoek2012practical,
  title={Practical bayesian optimization of machine learning algorithms},
  author={Snoek, Jasper and Larochelle, Hugo and Adams, Ryan P},
  journal={Advances in neural information processing systems},
  volume={25},
  year={2012}
}

@misc{wandb,
title = {Experiment Tracking with Weights and Biases},
year = {2020},
note = {Software available from wandb.com},
url={https://www.wandb.com/},
author = {Biewald, Lukas},
}

@misc{label-studio,
  title={{Label Studio}: Data labeling software},
  url={https://github.com/HumanSignal/label-studio},
  note={Open source software available from https://github.com/HumanSignal/label-studio},
  author={
    Maxim Tkachenko and
    Mikhail Malyuk and
    Andrey Holmanyuk and
    Nikolai Liubimov},
  year={2020-2025},
}

@inproceedings{khattab2024dspy,
  title={DSPy: Compiling Declarative Language Model Calls into Self-Improving Pipelines},
  author={Khattab, Omar and Singhvi, Arnav and Maheshwari, Paridhi and Zhang, Zhiyuan and Santhanam, Keshav and Vardhamanan, Sri and Haq, Saiful and Sharma, Ashutosh and Joshi, Thomas T. and Moazam, Hanna and Miller, Heather and Zaharia, Matei and Potts, Christopher},
  journal={The Twelfth International Conference on Learning Representations},
  year={2024}
}

@article{khattab2022demonstrate,
  title={Demonstrate-Search-Predict: Composing Retrieval and Language Models for Knowledge-Intensive {NLP}},
  author={Khattab, Omar and Santhanam, Keshav and Li, Xiang Lisa and Hall, David and Liang, Percy and Potts, Christopher and Zaharia, Matei},
  journal={arXiv preprint arXiv:2212.14024},
  year={2022}
}

@misc{qwen3technicalreport,
      title={Qwen3 Technical Report}, 
      author={Qwen Team},
      year={2025},
      eprint={2505.09388},
      archivePrefix={arXiv},
      primaryClass={cs.CL},
      url={https://arxiv.org/abs/2505.09388}, 
}

@inproceedings{kwon2023efficient,
  title={Efficient Memory Management for Large Language Model Serving with PagedAttention},
  author={Woosuk Kwon and Zhuohan Li and Siyuan Zhuang and Ying Sheng and Lianmin Zheng and Cody Hao Yu and Joseph E. Gonzalez and Hao Zhang and Ion Stoica},
  booktitle={Proceedings of the ACM SIGOPS 29th Symposium on Operating Systems Principles},
  year={2023}
}

@inproceedings{cardellino-etal-2017-legal,
    title = "Legal {NERC} with ontologies, {W}ikipedia and curriculum learning",
    author = "Cardellino, Cristian  and
      Teruel, Milagro  and
      Alonso Alemany, Laura  and
      Villata, Serena",
    editor = "Lapata, Mirella  and
      Blunsom, Phil  and
      Koller, Alexander",
    booktitle = "Proceedings of the 15th Conference of the {E}uropean Chapter of the Association for Computational Linguistics: Volume 2, Short Papers",
    month = apr,
    year = "2017",
    address = "Valencia, Spain",
    publisher = "Association for Computational Linguistics",
    url = "https://aclanthology.org/E17-2041/",
    pages = "254--259"
}

@inproceedings{leitner-etal-2020-dataset,
    title = "A Dataset of {G}erman Legal Documents for Named Entity Recognition",
    author = "Leitner, Elena  and
      Rehm, Georg  and
      Moreno-Schneider, Julian",
    editor = "Calzolari, Nicoletta  and
      B{\'e}chet, Fr{\'e}d{\'e}ric  and
      Blache, Philippe  and
      Choukri, Khalid  and
      Cieri, Christopher  and
      Declerck, Thierry  and
      Goggi, Sara  and
      Isahara, Hitoshi  and
      Maegaard, Bente  and
      Mariani, Joseph  and
      Mazo, H{\'e}l{\`e}ne  and
      Moreno, Asuncion  and
      Odijk, Jan  and
      Piperidis, Stelios",
    booktitle = "Proceedings of the Twelfth Language Resources and Evaluation Conference",
    month = may,
    year = "2020",
    address = "Marseille, France",
    publisher = "European Language Resources Association",
    url = "https://aclanthology.org/2020.lrec-1.551/",
    pages = "4478--4485",
    language = "eng",
    ISBN = "979-10-95546-34-4"
}

@inproceedings{kalamkar-etal-2022-named,
    title = "Named Entity Recognition in {I}ndian court judgments",
    author = "Kalamkar, Prathamesh  and
      Agarwal, Astha  and
      Tiwari, Aman  and
      Gupta, Smita  and
      Karn, Saurabh  and
      Raghavan, Vivek",
    editor = "Aletras, Nikolaos  and
      Chalkidis, Ilias  and
      Barrett, Leslie  and
      Goanț{\u{a}}, C{\u{a}}t{\u{a}}lina  and
      Preoțiuc-Pietro, Daniel",
    booktitle = "Proceedings of the Natural Legal Language Processing Workshop 2022",
    month = dec,
    year = "2022",
    address = "Abu Dhabi, United Arab Emirates (Hybrid)",
    publisher = "Association for Computational Linguistics",
    url = "https://aclanthology.org/2022.nllp-1.15/",
    doi = "10.18653/v1/2022.nllp-1.15",
    pages = "184--193"
}

@inproceedings{dai-etal-2025-laiw,
    title = "{LA}i{W}: A {C}hinese Legal Large Language Models Benchmark",
    author = "Dai, Yongfu  and
      Feng, Duanyu  and
      Huang, Jimin  and
      Jia, Haochen  and
      Xie, Qianqian  and
      Zhang, Yifang  and
      Han, Weiguang  and
      Tian, Wei  and
      Wang, Hao",
    editor = "Rambow, Owen  and
      Wanner, Leo  and
      Apidianaki, Marianna  and
      Al-Khalifa, Hend  and
      Eugenio, Barbara Di  and
      Schockaert, Steven",
    booktitle = "Proceedings of the 31st International Conference on Computational Linguistics",
    month = jan,
    year = "2025",
    address = "Abu Dhabi, UAE",
    publisher = "Association for Computational Linguistics",
    url = "https://aclanthology.org/2025.coling-main.716/",
    pages = "10738--10766"
}

@inproceedings{tjong-kim-sang-2002-introduction,
    title = "Introduction to the {C}o{NLL}-2002 Shared Task: Language-Independent Named Entity Recognition",
    author = "Tjong Kim Sang, Erik F.",
    booktitle = "{COLING}-02: The 6th Conference on Natural Language Learning 2002 ({C}o{NLL}-2002)",
    year = "2002",
    url = "https://aclanthology.org/W02-2024/"
}

@inproceedings{tjong-kim-sang-de-meulder-2003-introduction,
    title = "Introduction to the {C}o{NLL}-2003 Shared Task: Language-Independent Named Entity Recognition",
    author = "Tjong Kim Sang, Erik F.  and
      De Meulder, Fien",
    booktitle = "Proceedings of the Seventh Conference on Natural Language Learning at {HLT}-{NAACL} 2003",
    year = "2003",
    url = "https://aclanthology.org/W03-0419/",
    pages = "142--147"
}

@inproceedings{palen-michel-etal-2021-seqscore,
    title = "{S}eq{S}core: Addressing Barriers to Reproducible Named Entity Recognition Evaluation",
    author = "Palen-Michel, Chester  and
      Holley, Nolan  and
      Lignos, Constantine",
    editor = "Gao, Yang  and
      Eger, Steffen  and
      Zhao, Wei  and
      Lertvittayakumjorn, Piyawat  and
      Fomicheva, Marina",
    booktitle = "Proceedings of the 2nd Workshop on Evaluation and Comparison of NLP Systems",
    month = nov,
    year = "2021",
    address = "Punta Cana, Dominican Republic",
    publisher = "Association for Computational Linguistics",
    url = "https://aclanthology.org/2021.eval4nlp-1.5/",
    doi = "10.18653/v1/2021.eval4nlp-1.5",
    pages = "40--50"
}

@inproceedings{lignos-etal-2023-improving,
    title = "Improving {NER} Research Workflows with {S}eq{S}core",
    author = "Lignos, Constantine  and
      Kruse, Maya  and
      Rueda, Andrew",
    editor = "Tan, Liling  and
      Milajevs, Dmitrijs  and
      Chauhan, Geeticka  and
      Gwinnup, Jeremy  and
      Rippeth, Elijah",
    booktitle = "Proceedings of the 3rd Workshop for Natural Language Processing Open Source Software (NLP-OSS 2023)",
    month = dec,
    year = "2023",
    address = "Singapore",
    publisher = "Association for Computational Linguistics",
    url = "https://aclanthology.org/2023.nlposs-1.17/",
    doi = "10.18653/v1/2023.nlposs-1.17",
    pages = "147--152"
}

@inproceedings{labrak-etal-2024-zero,
    title = "A Zero-shot and Few-shot Study of Instruction-Finetuned Large Language Models Applied to Clinical and Biomedical Tasks",
    author = "Labrak, Yanis  and
      Rouvier, Mickael  and
      Dufour, Richard",
    editor = "Calzolari, Nicoletta  and
      Kan, Min-Yen  and
      Hoste, Veronique  and
      Lenci, Alessandro  and
      Sakti, Sakriani  and
      Xue, Nianwen",
    booktitle = "Proceedings of the 2024 Joint International Conference on Computational Linguistics, Language Resources and Evaluation (LREC-COLING 2024)",
    month = may,
    year = "2024",
    address = "Torino, Italia",
    publisher = "ELRA and ICCL",
    url = "https://aclanthology.org/2024.lrec-main.185/",
    pages = "2049--2066"
}

@inproceedings{subedi-etal-2024-exploring,
    title = "Exploring the Potential of Large Language Models ({LLM}s) for Low-resource Languages: A Study on Named-Entity Recognition ({NER}) and Part-Of-Speech ({POS}) Tagging for {N}epali Language",
    author = "Subedi, Bipesh  and
      Regmi, Sunil  and
      Bal, Bal Krishna  and
      Acharya, Praveen",
    editor = "Calzolari, Nicoletta  and
      Kan, Min-Yen  and
      Hoste, Veronique  and
      Lenci, Alessandro  and
      Sakti, Sakriani  and
      Xue, Nianwen",
    booktitle = "Proceedings of the 2024 Joint International Conference on Computational Linguistics, Language Resources and Evaluation (LREC-COLING 2024)",
    month = may,
    year = "2024",
    address = "Torino, Italia",
    publisher = "ELRA and ICCL",
    url = "https://aclanthology.org/2024.lrec-main.611/",
    pages = "6974--6979"
}

@inproceedings{wang-etal-2025-gpt,
    title = "{GPT}-{NER}: Named Entity Recognition via Large Language Models",
    author = "Wang, Shuhe  and
      Sun, Xiaofei  and
      Li, Xiaoya  and
      Ouyang, Rongbin  and
      Wu, Fei  and
      Zhang, Tianwei  and
      Li, Jiwei  and
      Wang, Guoyin  and
      Guo, Chen",
    editor = "Chiruzzo, Luis  and
      Ritter, Alan  and
      Wang, Lu",
    booktitle = "Findings of the Association for Computational Linguistics: NAACL 2025",
    month = apr,
    year = "2025",
    address = "Albuquerque, New Mexico",
    publisher = "Association for Computational Linguistics",
    url = "https://aclanthology.org/2025.findings-naacl.239/",
    doi = "10.18653/v1/2025.findings-naacl.239",
    pages = "4257--4275",
    ISBN = "979-8-89176-195-7"
}

@inproceedings{devlin-etal-2019-bert,
    title = "{BERT}: Pre-training of Deep Bidirectional Transformers for Language Understanding",
    author = "Devlin, Jacob  and
      Chang, Ming-Wei  and
      Lee, Kenton  and
      Toutanova, Kristina",
    editor = "Burstein, Jill  and
      Doran, Christy  and
      Solorio, Thamar",
    booktitle = "Proceedings of the 2019 Conference of the North {A}merican Chapter of the Association for Computational Linguistics: Human Language Technologies, Volume 1 (Long and Short Papers)",
    month = jun,
    year = "2019",
    address = "Minneapolis, Minnesota",
    publisher = "Association for Computational Linguistics",
    url = "https://aclanthology.org/N19-1423/",
    doi = "10.18653/v1/N19-1423",
    pages = "4171--4186"
}

@inproceedings{opsahl-ong-etal-2024-optimizing,
    title = "Optimizing Instructions and Demonstrations for Multi-Stage Language Model Programs",
    author = "Opsahl-Ong, Krista  and
      Ryan, Michael J  and
      Purtell, Josh  and
      Broman, David  and
      Potts, Christopher  and
      Zaharia, Matei  and
      Khattab, Omar",
    editor = "Al-Onaizan, Yaser  and
      Bansal, Mohit  and
      Chen, Yun-Nung",
    booktitle = "Proceedings of the 2024 Conference on Empirical Methods in Natural Language Processing",
    month = nov,
    year = "2024",
    address = "Miami, Florida, USA",
    publisher = "Association for Computational Linguistics",
    url = "https://aclanthology.org/2024.emnlp-main.525/",
    doi = "10.18653/v1/2024.emnlp-main.525",
    pages = "9340--9366"
}

@misc{liu2019robertarobustlyoptimizedbert,
      title={RoBERTa: A Robustly Optimized BERT Pretraining Approach}, 
      author={Yinhan Liu and Myle Ott and Naman Goyal and Jingfei Du and Mandar Joshi and Danqi Chen and Omer Levy and Mike Lewis and Luke Zettlemoyer and Veselin Stoyanov},
      year={2019},
      eprint={1907.11692},
      archivePrefix={arXiv},
      primaryClass={cs.CL},
      url={https://arxiv.org/abs/1907.11692}, 
}

@inproceedings{au-etal-2022-e,
    title = "{E}-{NER} {---} An Annotated Named Entity Recognition Corpus of Legal Text",
    author = "Au, Ting Wai Terence  and
      Lampos, Vasileios  and
      Cox, Ingemar",
    editor = "Aletras, Nikolaos  and
      Chalkidis, Ilias  and
      Barrett, Leslie  and
      Goanț{\u{a}}, C{\u{a}}t{\u{a}}lina  and
      Preoțiuc-Pietro, Daniel",
    booktitle = "Proceedings of the Natural Legal Language Processing Workshop 2022",
    month = dec,
    year = "2022",
    address = "Abu Dhabi, United Arab Emirates (Hybrid)",
    publisher = "Association for Computational Linguistics",
    url = "https://aclanthology.org/2022.nllp-1.22/",
    doi = "10.18653/v1/2022.nllp-1.22",
    pages = "246--255"
}

\appendix

%
%

\section{Prompt Templates}
\label{appendix:prompts}

Each entity type uses a shared system prompt structure with
type-specific user prompts. We show the common system prompt
once (\S\ref{appendix:system-prompt}), followed by
representative user prompts for each category.

\subsection{Shared System Prompt}
\label{appendix:system-prompt}

\begin{tcolorbox}[colback=systembg,colframe=framecolor,
  boxrule=0.3pt,arc=1.5pt,left=3pt,right=3pt,
  top=2pt,bottom=2pt,breakable,
  fonttitle=\bfseries\scriptsize,title=System Prompt]
{\fontsize{6.5pt}{8pt}\selectfont
\begin{verbatim}
Your input fields are:
1. `mention` (str): the mention to extract the type 
from
2. `context` (str): the context surrounding the 
mention
3. `possible_tags` (list[str]): list of possible 
level-2 tags

Your output fields are:
1. `reasoning` (str)
2. `tag` (str): the type of mention. MUST BE ONE OF THE
   POSSIBLE TAGS PROVIDED.

All interactions will be structured in the following 
way, with the appropriate values filled in.

[[ ## mention ## ]]
{mention}

[[ ## context ## ]]
{context}

[[ ## possible_tags ## ]]
{possible_tags}

[[ ## reasoning ## ]]
{reasoning}

[[ ## tag ## ]]
{tag}

[[ ## completed ## ]]

In adhering to this structure, your objective is:
Extract contiguous tokens referring to members of
congress, titles, or simple names, if any, from a 
list of string tokens. Output a list of tokens.
\end{verbatim}
}
\end{tcolorbox}

\subsection{Abstraction}
\label{appendix:prompt-abstraction}

\begin{tcolorbox}[colback=userbg,colframe=framecolor,
  boxrule=0.3pt,arc=1.5pt,left=3pt,right=3pt,
  top=2pt,bottom=2pt,breakable,
  fonttitle=\bfseries\scriptsize,title=User Prompt]
{\fontsize{6.5pt}{8pt}\selectfont
\begin{verbatim}
[[ ## mention ## ]]
Children

[[ ## context ## ]]
2023 Mr . Cardin ( for himself and Ms . Stabenow )
introduced the following bill ; which was read twice
and referred to the Committee on Finance A BILL To
amend title XXI of the Social Security Act to prohibit
lifetime or annual limits on dental coverage under the
Children ' s Health Insurance Program , and to require
wraparound coverage of dental services for certain
children under such program . Be it enacted by the
Senate and House of Representatives of the United
States of America in Congress assembled , SECTION 1 .
SHORT TITLE . This Act may

[[ ## possible_tags ## ]]
["Doctrine", "Fund", "Infrastructure", "Misc",
 "Program", "Session", "Specification", "System"]

Respond with the corresponding output fields, starting
with the field [[ ## reasoning ## ]], then
[[ ## tag ## ]], and then ending with the marker for
[[ ## completed ## ]].
\end{verbatim}
}
\end{tcolorbox}

\subsection{Act}
\label{appendix:prompt-act}

\begin{tcolorbox}[colback=userbg,colframe=framecolor,
  boxrule=0.3pt,arc=1.5pt,left=3pt,right=3pt,
  top=2pt,bottom=2pt,breakable,
  fonttitle=\bfseries\scriptsize,title=User Prompt]
{\fontsize{6.5pt}{8pt}\selectfont
\begin{verbatim}
[[ ## mention ## ]]
Defending Domestic Produce Production Act of 2023

[[ ## context ## ]]
seasonal industries affected by antidumping or
countervailing duty investigations , and for other
purposes . Be it enacted by the Senate and House of
Representatives of the United States of America in
Congress assembled , SECTION 1 . SHORT TITLE . This
Act may be cited as the `` Defending Domestic Produce
Production Act of 2023 '' . SEC . 2 . DEFINITIONS .
( a ) Core Seasonal Industry . -- Section 771 of the
Tariff Act of 1930 ( 19 U.S.C. 1677 ) is amended by
adding at the end the following : `` ( 37

[[ ## possible_tags ## ]]
["Amendment", "PublicAct"]

Respond with the corresponding output fields, starting
with the field [[ ## reasoning ## ]], then
[[ ## tag ## ]], and then ending with the marker for
[[ ## completed ## ]].
\end{verbatim}
}
\end{tcolorbox}

\subsection{Class}
\label{appendix:prompt-class}

\begin{tcolorbox}[colback=userbg,colframe=framecolor,
  boxrule=0.3pt,arc=1.5pt,left=3pt,right=3pt,
  top=2pt,bottom=2pt,breakable,
  fonttitle=\bfseries\scriptsize,title=User Prompt]
{\fontsize{6.5pt}{8pt}\selectfont
\begin{verbatim}
[[ ## mention ## ]]
foster youth

[[ ## context ## ]]
Congress ] [ From the U.S. Government Publishing
Office ] [ S. 102 Introduced in ( IS ) ] 118th
CONGRESS 1st Session S. 102 To amend title IV of the
Social Security Act to establish a demonstration grant
program to provide emergency relief to foster youth
and improve pre-placement services offered by foster
care stabilization agencies , and for other purposes .
IN THE SENATE OF THE UNITED STATES January 26 , 2023
Mrs . Fischer ( for herself and Mr . Hickenlooper )
introduced the following bill ; which was read twice
and

[[ ## possible_tags ## ]]
["Non-ProtectedClass", "ProtectedClass"]

Respond with the corresponding output fields, starting
with the field [[ ## reasoning ## ]], then
[[ ## tag ## ]], and then ending with the marker for
[[ ## completed ## ]].
\end{verbatim}
}
\end{tcolorbox}

\subsection{Document}
\label{appendix:prompt-document}

\begin{tcolorbox}[colback=userbg,colframe=framecolor,
  boxrule=0.3pt,arc=1.5pt,left=3pt,right=3pt,
  top=2pt,bottom=2pt,breakable,
  fonttitle=\bfseries\scriptsize,title=User Prompt]
{\fontsize{6.5pt}{8pt}\selectfont
\begin{verbatim}
[[ ## mention ## ]]
S. 104

[[ ## context ## ]]
[ Congressional Bills 118th Congress ] [ From the
U.S. Government Publishing Office ] [ S. 104
Introduced in Senate ( IS ) ] 118th CONGRESS 1st
Session S. 104 To amend title VII of the Tariff Act
of 1930 to provide for the treatment of core seasonal
industries affected by antidumping or countervailing
duty investigations , and for other purposes . IN THE
SENATE OF

[[ ## possible_tags ## ]]
["Bill", "Code", "Parenthetical", "Reference",
 "Report"]

Respond with the corresponding output fields, starting
with the field [[ ## reasoning ## ]], then
[[ ## tag ## ]], and then ending with the marker for
[[ ## completed ## ]].
\end{verbatim}
}
\end{tcolorbox}

\subsection{Organization}
\label{appendix:prompt-organization}

\begin{tcolorbox}[colback=userbg,colframe=framecolor,
  boxrule=0.3pt,arc=1.5pt,left=3pt,right=3pt,
  top=2pt,bottom=2pt,breakable,
  fonttitle=\bfseries\scriptsize,title=User Prompt]
{\fontsize{6.5pt}{8pt}\selectfont
\begin{verbatim}
[[ ## mention ## ]]
Senate

[[ ## context ## ]]
Committee on Finance A BILL To amend title IV of the
Social Security Act to establish a demonstration grant
program to provide emergency relief to foster youth
and improve pre-placement services offered by foster
care stabilization agencies , and for other purposes .
Be it enacted by the Senate and House of
Representatives of the United States of America in
Congress assembled , SECTION 1 . SHORT TITLE . This
Act may be cited as the `` Foster Care Stabilization
Act of 2023 '' . SEC . 2 . GRANTS TO IMPROVE PRE -
PLACEMENT SERVICES FOR

[[ ## possible_tags ## ]]
["Agency", "Association", "Committee",
 "InternationalInstitution", "LegislativeBody",
 "Locality", "Nation", "State"]

Respond with the corresponding output fields, starting
with the field [[ ## reasoning ## ]], then
[[ ## tag ## ]], and then ending with the marker for
[[ ## completed ## ]].
\end{verbatim}
}
\end{tcolorbox}

\subsection{Person}
\label{appendix:prompt-person}

\begin{tcolorbox}[colback=userbg,colframe=framecolor,
  boxrule=0.3pt,arc=1.5pt,left=3pt,right=3pt,
  top=2pt,bottom=2pt,breakable,
  fonttitle=\bfseries\scriptsize,title=User Prompt]
{\fontsize{6.5pt}{8pt}\selectfont
\begin{verbatim}
[[ ## mention ## ]]
Secretary

[[ ## context ## ]]
this subsection shall have 3 years to spend funds
awarded by the grant and return any unused grant funds
to the Secretary . `` ( 3 ) Application . -- A foster
care stabilization agency that desires to receive a
grant under this subsection shall submit to the
Secretary an application at such time , in such
manner , and containing such information as the
Secretary may require , that shall include the
following : `` ( A ) A description of how grant funds
will be used to provide emergency relief to foster
youth by the foster

[[ ## possible_tags ## ]]
["Member", "Name", "Title"]

Respond with the corresponding output fields, starting
with the field [[ ## reasoning ## ]], then
[[ ## tag ## ]], and then ending with the marker for
[[ ## completed ## ]].
\end{verbatim}
}
\end{tcolorbox}

\renewcommand\tabularxcolumn[1]{m{#1}}

\section{Annotation Guidelines}

Our guidelines consist of a set of rules with examples as well as definitions for tags at both levels.

\begin{table*}[ht]
  \small
  \centering
    \begin{tabularx}{\textwidth}{|c|l|X|m{4.5cm}|}
    \hline
    \textbf{Top Tag} & \textbf{Subclass} & \textbf{Description} & \textbf{Examples} \\
    \hline
    \multirow{3}{*}[-1\normalbaselineskip]{\textbf{Person}} & Member \textbf{MBC} & A member of Congress, typically the sponsor or co-sponsor of the legislation. & Mr. \PER{Biggs}{MBC} \\
    \cline{2-4}
    & Title \textbf{TTL} & Mentions of individuals by their official position or office. & \PER{Speaker}{TTL}; \PER{President}{TTL} \\
    \cline{2-4}
    & Individual \textbf{IND} & Any specific named person who is neither a Member of Congress nor a Title. & \PER{John Doe}{IND} \\
    \hline
  \end{tabularx}
  \caption{Tag Definitions: \textsc{Person} Subclasses}
  \label{tab:person-ontology}
\end{table*}

\begin{table*}[ht]
  \small
  \centering
  \begin{tabularx}{\textwidth}{|c|l|X|p{4.5cm}|}
    \hline
    \textbf{Top Tag} & \textbf{Subclass} & \textbf{Description} & \textbf{Examples} \\
    \hline
    \multirow{8}{*}[-2.5\normalbaselineskip]{\textbf{Organization}} & Nation \textbf{NAT} & National-level geopolitical entities (includes American Indian nations). & \ORG{United States}{NAT} \\
    \cline{2-4}
    & State \textbf{STA} & State-level geopolitical entities within the U.S. & \ORG{Illinois}{STA} \\
    \cline{2-4}
    & Locality \textbf{LOC} & Local geopolitical entities, such as cities, counties, or municipalities. & \ORG{Cook County}{LOC} \\
    \cline{2-4}
    & Committee \textbf{COM} & Named Congressional committees and subcommittees. & \ORG{Ways and Means}{COM} \\
    \cline{2-4}
    & Agency \textbf{AGC} & Federal departments, bureaus, or executive administrations. & \ORG{Dept. of Justice}{AGC} \\
    \cline{2-4}
    & Leg. Body \textbf{LEG} & Primary legislative chambers (Senate, House, Congress). & \ORG{Senate}{LEG} \\
    \cline{2-4}
    & International \textbf{INT} & International organizations and multi-national institutions. & \ORG{United Nations}{INT} \\
    \cline{2-4}
    & Association \textbf{ASC} & Catch-all for other named organizations, nonprofits, or private entities. & \ORG{Planned Parenthood}{ASC} \\
    \hline
  \end{tabularx}
  \caption{Tag Definitions: \textsc{Organization} Subclasses}
  \label{tab:org-ontology}
\end{table*}

\begin{table*}[ht]
  \small
  \centering
  \begin{tabularx}{\textwidth}{|c|l|X|p{4.5cm}|}
    \hline
    \textbf{Top Tag} & \textbf{Subclass} & \textbf{Description} & \textbf{Examples} \\
    \hline
    \multirow{6}{*}[-2\normalbaselineskip]{\textbf{Document}} & Code \textbf{CODE} & Foundational legal codes and structured statutory bodies. & \DOC{U.S. Code}{CODE} \\
    \cline{2-4}
    & Bill \textbf{BILL} & Specific bill identifiers and tracking numbers. & \DOC{H.R. 189}{BILL} \\
    \cline{2-4}
    & Reference \textbf{REF} & Structural subdivisions of a legal text (sections, titles, paragraphs). & \DOC{section 102}{REF} \\
    \cline{2-4}
    & Parenthetical \textbf{PAR} & Legal citations provided in parenthetical format. & \DOC{(42 U.S.C. 4332)}{PAR} \\
    \cline{2-4}
    & Treaty \textbf{TRE} & Named international treaties and conventions. & \DOC{Geneva Convention}{TRE} \\
    \cline{2-4}
    & Report \textbf{REP} & Official reports, budget documents, or findings. & \DOC{President's Budget}{REP} \\
    \hline
  \end{tabularx}
  \caption{Tag Definitions: \textsc{Document} Subclasses}
  \label{tab:doc-ontology}
\end{table*}

\begin{table*}[ht]
  \small
  \centering
  \begin{tabularx}{\textwidth}{|c|l|X|p{4.5cm}|}
    \hline
    \textbf{Top Tag} & \textbf{Subclass} & \textbf{Description} & \textbf{Examples} \\
    \hline
    \multirow{2}{*}[-0.5\normalbaselineskip]{\textbf{Act}} & Public Act \textbf{PA} & Named acts or legislation referred to by their popular title. & \ACT{Green New Deal}{PA} \\
    \cline{2-4}
    & Amendment \textbf{AMD} & Specific amendments to the U.S. Constitution. & \ACT{Second Amendment}{AMD} \\
    \hline
  \end{tabularx}
  \caption{Tag Definitions: \textsc{Act} Subclasses}
  \label{tab:act-ontology}
\end{table*}

\begin{table*}[ht]
  \small
  \centering
  \begin{tabularx}{\textwidth}{|c|l|X|p{4.5cm}|}
    \hline
    \textbf{Top Tag} & \textbf{Subclass} & \textbf{Description} & \textbf{Examples} \\
    \hline
    \multirow{8}{*}[-2.5\normalbaselineskip]{\textbf{Abstraction}} & Program \textbf{PRO} & Named government-managed programs and initiatives. & \ABS{Medicare}{PRO} \\
    \cline{2-4}
    & System \textbf{SYS} & Intangible government frameworks, databases, or digital networks. & \ABS{Forest System}{SYS} \\
    \cline{2-4}
    & Infrastructure \textbf{INF} & Tangible or physical frameworks established by government action. & \ABS{Interceptor}{INF} \\
    \cline{2-4}
    & Fund \textbf{FUND} & Specific, named accounts or monetary trust funds. & \ABS{General Fund}{FUND} \\
    \cline{2-4}
    & Doctrine \textbf{DTR} & Abstract legal policies, principles, or constitutional rights. & \ABS{Free Speech}{DTR} \\
    \cline{2-4}
    & Case \textbf{CASE} & Court cases and judicial precedents. & \ABS{McCulloch v. MD}{CASE} \\
    \cline{2-4}
    & Specification \textbf{SPC} & Specific legal classifications, modifiers, or schedules. & \ABS{340B}{SPC} \\
    \cline{2-4}
    & Session \textbf{SES} & Named intervals, meetings, or sessions of a legislative body. & \ABS{118th Congress}{SES} \\
    \hline
  \end{tabularx}
  \caption{Tag Definitions: \textsc{Abstraction} Subclasses}
  \label{tab:abs-ontology}
\end{table*}

\begin{table*}[ht]
  \small
  \centering
  \begin{tabularx}{\textwidth}{|c|l|X|p{4.5cm}|}
    \hline
    \textbf{Top Tag} & \textbf{Subclass} & \textbf{Description} & \textbf{Examples} \\
    \hline
    \multirow{2}{*}[-0.5\normalbaselineskip]{\textbf{Class}} & Protected \textbf{PC} & Groups defined by law for special protections (Race, Disability, etc.). & \CLS{veterans}{PC} \\
    \cline{2-4}
    & Non-Prot. \textbf{NPC} & Other groups that are the subject of legislation but not protected. & \CLS{refugees}{NPC} \\
    \hline
  \end{tabularx}
  \caption{Tag Definitions: \textsc{Class} Subclasses}
  \label{tab:class-ontology}
\end{table*}

\subsection{Rules}
\begin{enumerate}
    \item No Nesting \begin{enumerate}
        \item A token may be part of one tag at maximum
        \item A token should be part of the largest possible tag and only that tag
    \end{enumerate}
    
\resizebox{.45\textwidth}{!}{
\begin{tabular}{ll}

    \PER{Secretary of Veterans’ Affairs}{TTL} & YES\\
    \PER{Secretary}{TTL} of \\
    \ORG{Veterans’ Affairs}{AGC} & NO\\
    \PERP{\PER{Secretary}{TTL} of }{} \\
    \PER{\ORG{Veterans’ Affairs}{AGC}}{TTL} & NO
\end{tabular}
}

    \item Person tags do not include titles
    
\begin{tabular}{l}
    Mr. \PER{Biggs}{MBC} \\
    Administrator \PER{Sanders}{IND} \\
\end{tabular}

    \item Person tags include shortened named references 
    
\begin{tabular}{l}
    the \PER{Speaker}{TTL} \\
\end{tabular}

    \item ONLY use Document tags outside of quotations
    
\begin{tabular}{l}
    Amend \DOC{Section 1710F}{REF} as, “1710F: …” \\
\end{tabular}

    \item ONLY use Document tags if the Act name or full reference is mentioned in the text, not when only a section or paragraph is mentioned
\resizebox{.45\textwidth}{!}{
\begin{tabular}{lc}
    \DOCP{paragraph one of the twenty-seventh}{} & \\
    \DOCP{article of amendment to the Constitution}{} & \\
    \DOC{of the United States}{REF} & YES \\
    \DOC{section (1)(b)}{REF} & NO \\
    \DOC{section (1)(b) of US Code3 …}{REF} & YES
\end{tabular}
}

    \item ONLY Document tags span over “and” and “or”
\resizebox{.45\textwidth}{!}{
\begin{tabular}{lc}
    \DOC{sections (a) and (b) of US Code  ….}{REF} & YES \\
    \PER{John and Michael}{IND} & NO \\
    \PER{John}{IND} and \PER{Michael}{IND} & YES
\end{tabular}
}

    \item Administrative actions in bills, including “Clerical Amendment”, and “Date of Effect” should not be tagged

    \item If any information below the Act title is included, tag the mention as Document (Title, section, paragraph, part, etc.)
\resizebox{.45\textwidth}{!}{
\begin{tabular}{l}
    \DOCP{Title IX of the Farm Security and Rural}{} \\
    \DOC{Investment Act of 2002}{REF} \\
    \ACTP{Farm Security and Rural Investment}{} \\
    \ACT{Act of 2002}{PA} \\
\end{tabular}
}

    \item Tag parenthesized mentions of Documents separately
    
\resizebox{.45\textwidth}{!}{
\begin{tabular}{l}
    \DOCP{Title IX of the Farm Security and Rural}{} \\
    \DOC{Investment Act of 2002}{REF} \DOCP{(7 U.S.C.}{} \\
    \DOC{8101 et seq.)}{PAR}\\
\end{tabular}
}

    \item Do not tag dates, currency, or any other numbers

\end{enumerate}

\clearpage

\begin{table*}
    \section{Summary Statistics}
    \centering
    \begin{tabular}{ll|lll|l}
        \toprule
        \textbf{Type} & & \textbf{Train} & \textbf{Dev} & \textbf{Test} & \textbf{Total}  \\
        \midrule
        Abstraction                       & & 217   & 91    & 79    & 387   \\               
            & Case                          & 0     & 0     & 0     & 0     \\               
            & Doctrine                      & 8     & 0     & 1     & 9     \\               
            & Fund                          & 39    & 11    & 1     & 51    \\               
            & Infrastructure                & 3     & 0     & 21    & 24    \\        
            & Misc                          & 4     & 0     & 0     & 4     \\        
            & Program                       & 30    & 39    & 18    & 87    \\           
            & Session                       & 87    & 21    & 31    & 139   \\           
            & Specification                 & 14    & 17    & 0     & 31    \\         
            & System                        & 32    & 3     & 7     & 41    \\                
            \midrule
        Act                               & & 197   & 40    & 80    & 317   \\                           
            & Amendment                     & 3     & 1     & 1     & 5     \\                     
            & Public Act                    & 194   & 39    & 79    & 312   \\   
            \midrule
        Class                             & & 208   & 57    & 67    & 332   \\                     
            & Non-Protected Class           & 194   & 45    & 59    & 298   \\   
            & Protected Class               & 14    & 12    & 8     & 34    \\        
            \midrule
        Document                          & & 774   & 214   & 350   & 1,338 \\                              
            & Bill                          & 162   & 42    & 60    & 264   \\             
            & Code                          & 30    & 2     & 5     & 37    \\
            & Parenthetical                 & 204   & 32    & 112   & 348   \\
            & Reference                     & 361   & 131   & 171   & 663   \\
            & Report                        & 17    & 7     & 2     & 26    \\
            & Treaty                        & 0     & 0     & 0     & 0     \\
            \midrule
        Organization                      & & 1261  & 379   & 579   & 2,219 \\
            & Agency                        & 249   & 78    & 191   & 518   \\
            & Association                   & 28    & 38    & 38    & 104   \\
            & Committee                     & 158   & 44    & 78    & 280   \\
            & International Institution     & 1     & 0     & 3     & 4     \\
            & Legislative Body              & 566   & 150   & 194   & 910   \\
            & Locality                      & 20    & 7     & 28    & 55    \\
            & Nation                        & 178   & 35    & 120   & 333   \\
            & State                         & 61    & 27    & 17    & 105   \\
            \midrule
        Person                            & & 731   & 248   & 238   & 1,217 \\
            & Member                        & 339   & 106   & 121   & 566   \\
            & Name                          & 4     & 0     & 0     & 4     \\
            & Title                         & 388   & 142   & 117   & 647   \\
        \bottomrule
    \end{tabular}
    \caption{Counts of total level-one and level-two mentions by split}
    \label{tab:counts}
\end{table*}

\end{document}